\documentclass[10pt,twocolumn,letterpaper,amsart]{article}

\usepackage[pagenumbers]{cvpr} 

\usepackage{graphicx}
\usepackage{amsmath}
\usepackage{amssymb}
\usepackage{booktabs}
\usepackage{enumitem} 
\usepackage{xcolor}
\usepackage{tabularx}
\usepackage{multicol, blindtext}

\usepackage[pagebackref,breaklinks,colorlinks]{hyperref}

\usepackage[capitalize]{cleveref}
\crefname{section}{Sec.}{Secs.}
\Crefname{section}{Section}{Sections}
\Crefname{table}{Table}{Tables}
\crefname{table}{Tab.}{Tabs.}

\def\cvprPaperID{*****} 
\def\confName{CVPR}
\def\confYear{2022}

\begin{document}

\title{ MultiEarth 2022 Deforestation Challenge - ForestGump }

\author{Dongoo Lee\thanks{Equal contribution} \quad Yeonju Choi\footnotemark[1]  \thanks{Corresponding author}\\
 Korea Aerospace Research Institute\\
{\tt\small 	\{ldg810, choiyj\}@kari.re.kr}
}
\maketitle

\begin{abstract}
The estimation of deforestation in the Amazon Forest is challenge task because of the vast size of the area and the difficulty of direct human access. However, it is a crucial problem in that deforestation results in serious environmental problems such as global climate change, reduced biodiversity, etc. In order to effectively solve the problems, satellite imagery would be a good alternative to estimate the deforestation of the Amazon. With a combination of optical images and Synthetic aperture radar (SAR) images, observation of such a massive area regardless of weather conditions become possible. In this paper, we present an accurate deforestation estimation method with conventional U-Net and comprehensive data processing. The diverse channels of Sentinel-1, Sentinel-2 and Landsat 8 are carefully selected and utilized to train deep neural networks. With the proposed method, deforestation status for novel queries are successfully estimated with high accuracy.
\end{abstract}

\section{Introduction}
\label{sec:intro}
The Amazon accounts for about 40\% of the global rainforest, and the deforestation of the Amazon rainforest, which is called the 'lungs of the earth', is approaching the worst in history. Tropical rainforests are being destroyed by cultivation for ranching, indiscriminate fires and illegal logging. The area of deforestation, which was 98,000 km$^{2}$ until the 1970s, has increased tenfold in 40 years and the rate of destruction is increasing every year \cite{de2009estimativa}. According to the results announced by the Brazilian National Space Research Institute (INPE) based on satellite image monitoring, the area of deforestation in January 2022 recorded the highest ever recorded \cite{news_brazil}. INPE has been monitoring deforestation using satellite data since the 1980s, and it is a system to detect illegal logging early and respond before damage becomes large (PRODES (Brazilian Amazon Rainforest Monitoring Program by Satellite) and the DETER (Real-time Deforestation Detection System \cite{de2020change}).  

Remote sensing data enables real-time monitoring of a wide area, which is very effective in detecting deforestation. Traditional image-based deforestation detection methods largely used satellite image bands synthesis and threshold-based methods, but as deep learning has lately been used to remote sensing, machine learning-based deforestation detection research is on the rise. De Bem\cite{de2009estimativa} evaluated various machine learning techniques for deforestation estimation in Brazil based on Landsat-8, and showed that FCN (Fully Connected Network) is suitable for deforestation compared to other algorithms. And Isaienkov \etal \cite{isaienkov2020deep} suggested the change detection method in the Ukrainian forest using a segmentation network (U-net) and Sentinel-2 data. An attention U-net segmentation network using Sentinel-2 is proposed for deforestation detection in South America, the Amazon Rainforest and the Atlantic\cite{john2022attention}. In addition, using Land8 and Sen2 data, a study to predict the region where deforestation occurs in two matched images with time intervals has been introduced \cite{torres2021deforestation}. Torres \etal \cite{torres2021deforestation} applied a weighted cross-entropy loss to solve the deforestation and class imbalance problem in the surrounding region, and ResU-Net and FC DenseNet showed excellent performance. On the other hand, there are few studies of deforestation on a global scale, and most of these studies target a local specific area. Masolele \etal. \cite{masolele2021spatial} analyzed the difference in detection accuracy by continent, proposed a global model, and inferred spatial bias through this.

The remainder of this paper is organized as follows. \cref{sec:data} introduced the contents and generation method of the dataset in detail and \cref{sec:metho} described the proposed methodology for deforestation detection and in \cref{sec:res}, prediction results are presented. The conclusion is presented in \cref{sec:con}.

\section{Dataset}
\label{sec:data}

\begin{figure}
  \centering
  \includegraphics[height=50mm, width=75mm]{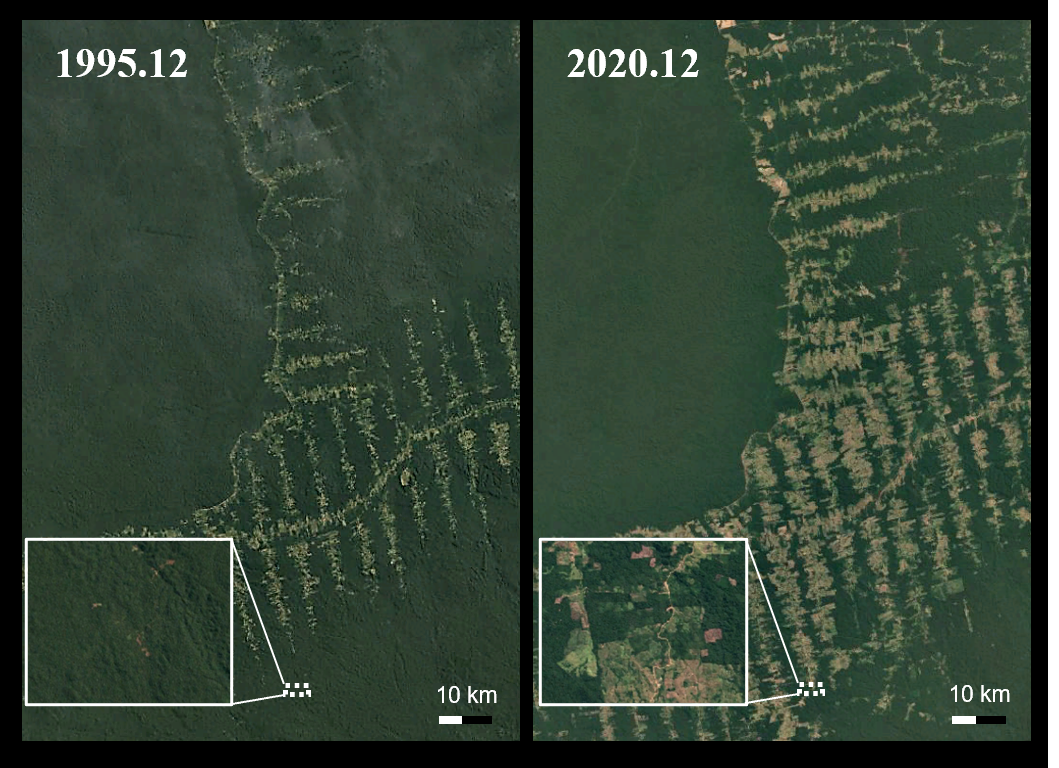}
  \caption{Region of interest and example of deforestation area in white box.}
  \label{fig:roi}
\end{figure}

The region of interest in this study is a part of the Amazon (latitude (LAT) :-3.33$^{\circ}$ $\sim$ -4.39$^{\circ}$ , longitude (LON) :-54.48$^{\circ}$ $\sim$ 55.2$^{\circ}$) as shown in \cref{fig:roi}, where deforestation has progressed rapidly in 30 years\cite{googleearth}.

\begin{figure}[h] 
  \centering
  \includegraphics{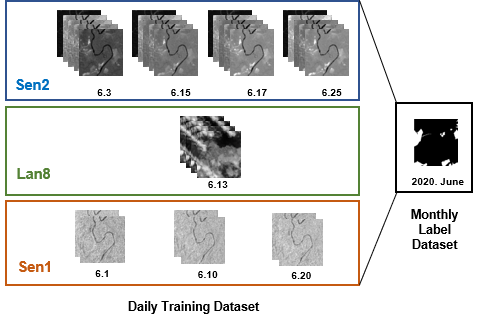}
  \caption{Example of pair with training  and labeling data.}
  \label{fig:dataset}
\end{figure}

The multimodal remotes sensing dataset provided by the Multiearth competition\cite{https://doi.org/10.48550/arxiv.2204.07649} to estimate the deforestation of the region consists of Sentinel-1 (Sen1), Sentinel-2 (Sen2), Landsat 5 (Land5), and Landsat 8 (Land8). There is a difference in date, bands, and resolution between each satellite. Sen1 is a synthetic aperture radar (SAR) satellite, which has two polarization bands: VV (vertical transmit/vertical receive) and VH (vertical transmit/horizontal receive) with 10 m spatial resolution. Sen2 has 13 spectral bands that range from the visible to the shortwave infrared (SWIR) with variable resolutions from 10 to 60 m. The Land5 and Land8  images consist of 7 spectral bands with 30 m spatial resolution and 11 spectral bands with variable resolutions from 150 to 100 m, respectively. 
As described in \cite{https://doi.org/10.48550/arxiv.2204.07649},the provided dataset consists of total 32 bands from 4 types of satellites, including the cloud quality layer. Therefore, it was required to select the relevant band and match the resolution from a large data set. First, to unified the resolution of satellites, Land8, which has a lower resolution than Sen1,2, was resized to 256 $\times$ 256 with 10 m resolution. To the next, it was confirmed that the deforestation labeling data only contains data for the year from 2016 $\sim$ 2021, therefore, we assumed that Land5, which contains only images collected until 2012, would not have affected the generation of the labeling data. Therefore, the Land5 images were excluded from the training set as data cleanings process. 
Meanwhile, raw satellite image is collected daily (but not continuously), but the labeling data has only one image representing each month.  We assumed that all raw data for a given month is used to predict one labeling image. As shown in \cref{fig:dataset}, a training pair set was created by matching daily satellite data to one labeling data. These dataset were directly used for training to predict the deforestation, it means that the change detection method was not applied on this approach.

In the deforestation classification study, deforestation region were classified by analyzing the reflectance of satellite images that appear differently in various regions such as forests, water systems, and cities \cite{candra2020deforestation,lima2019comparing}. According to these research, We selected Normalized Burn Ratio (NBR) and Normalized Difference Vegetation Index (NDVI) to detect the deforested region. The NBR described in \cref{eq:nbr} is a ratio-based vegetation index, and a value closer to 1 means closed forest cover, and a smaller or negative number means opening region. In addition, the NDVI in \cref{eq:ndvi} is used as one of the greenness indicators of vegetation.

\begin{equation}
  NBR = \frac{NIR-SWIR}{NIR + SWIR}
  \label{eq:nbr}
\end{equation}

\begin{equation}
  NDVI = \frac{NIR - RED}{NIR + RED }
  \label{eq:ndvi}
\end{equation}

Based on the above indices, we selected a total 11 bands for this study and the bands for each satellite finally selected as the training dataset are described in \cref{tab:seldata}.

\begin{table*}[ht]
\centering
\begin{tabular}{cccc}
\hline   \\[-1ex]
\textbf{Satellite}  & \textbf{Time} & \textbf{Selected Bands} & \textbf{Size}  \\  
\hline  & \\[-1ex]
\textbf{Sentinel-1} & 2014-2017     & VV, VH & (2,256,256)   \\ [10pt]
\textbf{Sentinel-2} & 2018-2021     &
\begin{tabular}[c]{@{}c@{}}B4 (RED), B7 (Vegetation red edge),
\\ B8 (NIR), B11 (SWIR 1), B12 (SWIR 2)
\end{tabular} & (5,256,256)   \\ [10pt]
\textbf{Landsat 8}  & 2013-2021     & B4 (RED), B5 (NIR),B6 (SWIR 1), B7 (SWIR 2)                       
& (4,256,256)   \\[10pt]
\textbf{Labeling}   & 2016-2021     & Binary {[}0,1{]} & (2,256,256)   \\  [10pt]    
\hline
\end{tabular}
\caption{Selected satellite and bands for the training set.}
\label{tab:seldata}
\end{table*}

The number of training images finally included in each satellite dataset was 70,976 for Sen1, 69,350 for Sen2 and 33,175 for Land8.

\section{Methodology}
\label{sec:metho}
\begin{figure}[h]
  \centering
  \includegraphics[width=80mm,height=35mm]{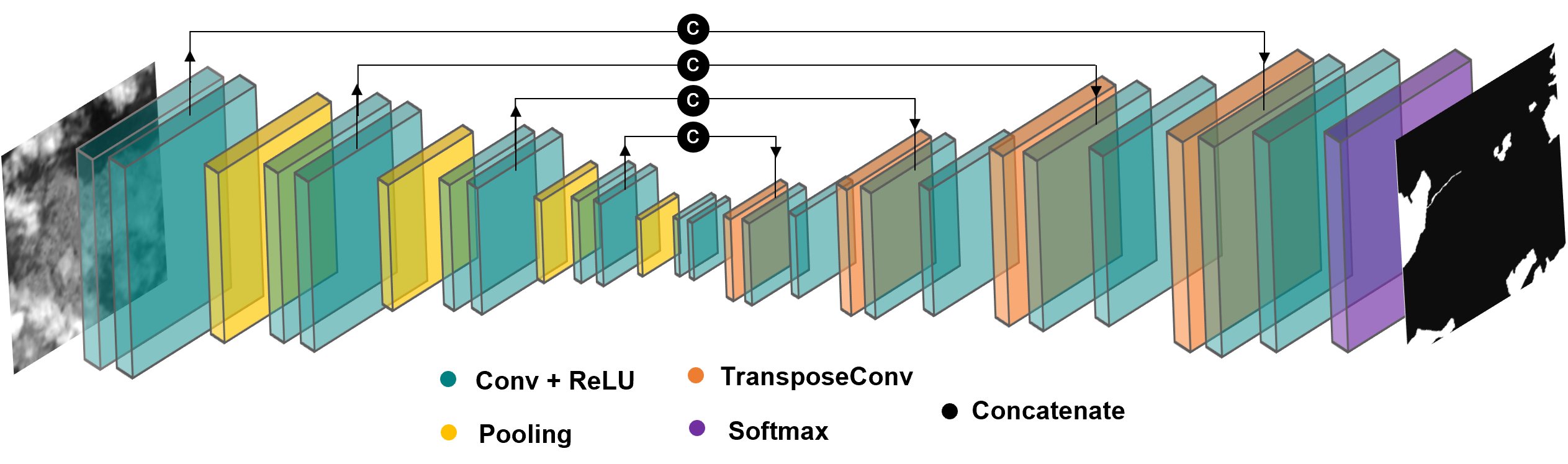}
  \caption{A network architecture diagram for U-Net.}
  \label{fig:u-net}
\end{figure}

The deforestation estimation task is a binary segmentation problem to determine whether a region is deforested or not. To solve the problem, we adopted conventional U-Net\cite{ronneberger2015u} as depicted in \cref{fig:u-net} which is known to be effective for segmentation problems in satellite imagery \cite{kislov2021extending}. With a conventional deep neural network, we focused on data pre-processing and post-processing to improve performance. The overall procedures for deforestation estimation method is as follows.

\subsection{Pre-processing}
\label{sec:preprocessing}

The given dataset comprises an excessive number of bands and some bands may not be relevant to deforestation detection. As mentioned in the previous section, we carefully selected relevant bands for Sen2 and Land8 imagery data. The pre-processing procedures of training dataset are summarized as follows.
\begin{enumerate}[label=\roman*]
\item Select relevant bands for deforestation detection as described in \cref{tab:seldata}
\item Generate a training input image with only selected bands. The shape of the processed images are (2, 256, 256) for Sen1, (5, 256, 256) for Sen2 and (4, 256, 256) for Land8.
\item All three types of training datasets were normalized to [0,1].
\end{enumerate}

\subsection{Training Network}
\label{sec:trainnet}
To detect the deforestation in this study, we took pairs of co-registered train dataset and labeling data as described in \cref{sec:data}. In this study, there are two detection classes, the deforested area, which is the target, is class 1, and the forested/other area is class 0. To compensate for the class imbalance, we adopted the combination of  Binary Cross-Entropy loss\cite{yi2004automated} and the Dice loss\cite{sudre2017generalised}. In addition, the model training is performed with a batch size of 16, RMSProp optimization, and the learning rate is adjusted by checking for validation loss every two epochs.

\subsection{Post processing}
\label{sec:post}
The most difficult point in this challenge was that the test queries are discretized as month with a nominal day label of particular regions, expressed in latitude and longitude. However, there are many cases for each query such as only one kind of satellite image is available and two or three kinds of satellite images are available for the a given query. Participants should merge the various information to one output image as a query requested. We did the following procedures to make final deforestation estimation output for the given queries.

\begin{figure*}[ht]
\centering
\includegraphics[width=170mm]{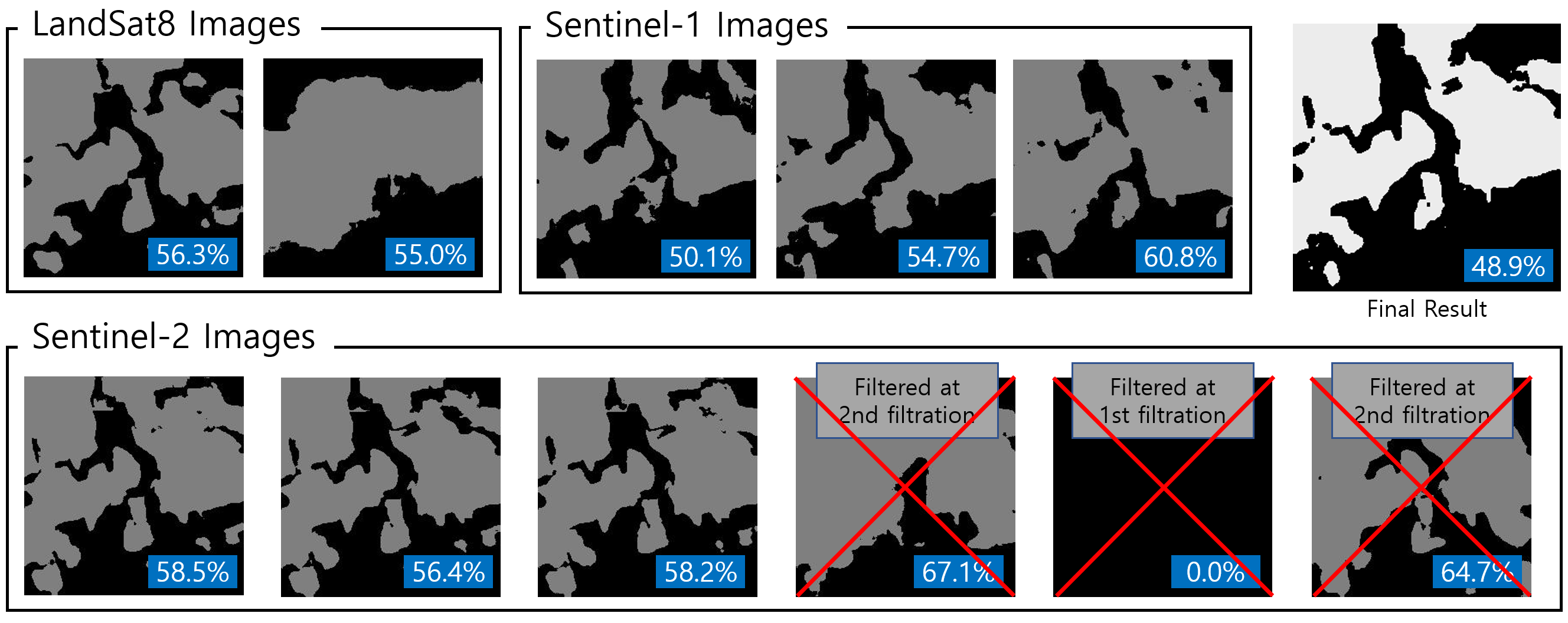}
\caption{Example images for a test query(LAT:-3.67, LON:-54.80, Date: August 2020).}
\label{fig:examplequery}
\end{figure*}

\begin{enumerate}[label=\roman*]
\item  Generate all of the possible outputs from images within the month of a specific query using three identical network that differ only in input size (Sen1, Sen2 and Land8).
\item Clustering the prediction results by the same month. Despite learning from the same network, some of the prediction results were completely black (no deforestation), and some present plausible deforestation results. Therefore, two-step filtering was applied to the output result to ensure the overall detection performance. The clear outliers were filtered without three-sigma range ($\mu$ $\pm$ 3$\sigma$) at the first filtration step. Second filtration used one-sigma range  ($\mu$ $\pm$ $\sigma$) using predicted deforestation percentage in the images. Here, $\mu$ is the mean and $\sigma$ is the variation of deforestation ratio for the predicted output.
\item  With filtered output images in the previous step, all images are averaged into the one binary image and each pixel value is assigned to 1 if the probability of the deforestation in the pixel is over certain threshold value (we set the value as 40\%). If the probability is lower than the threshold, pixel value was assigned to 0.
\item  Finally, the noise in binary image is removed  with the morphology opening operation $\ominus$ as described in \cref{eq:opening}  that is dilation operation of erosion result.

\begin{equation}
    I\circ M  = (I \ominus M)\oplus M,
     \label{eq:opening}
\end {equation}
where, \textit{I}, \textit{M} are the original image and a structuring element and $\ominus$, $\oplus$ are the erosion and dilation operation respectively. The erosion  operation eliminates small objects and dilation restores the size and shape of the remaining objects in image.

\end{enumerate}

\section{Results}
\label{sec:res}

The test set consists of 1,000 queries within August 2016 to August 2021 for 119 regions. There are at least one image available for each query and 14 images are available for the most informative case.

For an example result, we visualize whole processes for a test query of (LAT:-3.67,LON:-54.80,Date:August 2020) in \cref{fig:examplequery}. There are a total of 11 images of different dates/satellites available for the month(2 for Land8, 3 for Sen1 and 6 for Sen2). With three trained networks, 11 predicted images are generated. The range of deforestation percentage for the predictions are 0.0\% $\sim$ 67.1\%. The deforestation ratio of 0.0\% which is obviously outlier among the prediction results, was excluded at the first filtration step with three-sigma criterion. After the step, with 10 remaining results, two images with deforestation ratio of 64.7\% and 67.1\% are also excluded in second filtration step with one-sigma criterion. Finally, integrated result for the query is made with 8 out of 11 images with averaging and denoising process.

\begin{table}[h]
\centering
\begin{tabular}{ccc} 
\hline \\[-2ex]
\textbf{Pixel Accuracy} & \textbf{F1 Score} & \textbf{IoU}    \\ \hline \\[-2ex]
91.7102        & 0.8394   & 0.7644 \\ \hline
\end{tabular}
\caption{Final statistics for test set evaluation.}
\label{tab:result}
\end{table}

The proposed method finally achieves 91.71 pixel accuracy, 0.84 F1-score and 0.76 IoU for the test set as represented in \cref{tab:result}. With filtration and denoising in the post-processing procedure as described in \cref{sec:post}, final detection performance has been improved from the initial results.

\section{Conclusion}
\label{sec:con}

Satellite imagery is a very efficient solution to estimate deforestation of the Amazon rainforest area since the area of interest is very spacious. In the MultiEarth 2022 challenge, four kinds of satellite imagery (Sentinel-1, Sentinel-2, Landsat 5, Landsat 8) are provided to estimate deforestation. We carefully selected 11 channels (Sentinel-1 :VV,VH, Sentinel-2: B4,B7,B8,B11,B12, Landsat 8: B4,B5,B6,B7) based on the spectral feature of the satellite band in the deforestation area. Because there is a lot of data that is different for each satellite and date, we trained three identical neural networks for each satellite type. The three networks are used to predict deforestation and results with the same date range and location merged into a final image excluding outliers and noise. High performance was achieved through proposed post processing steps, which was applied to transform multiple prediction results into a single representative result.

{\small
\bibliographystyle{ieee_fullname}
\bibliography{egbib}
}

\end{document}